\documentclass[sigconf]{acmart}

\AtBeginDocument{%
  \providecommand\BibTeX{{%
    \normalfont B\kern-0.5em{\scshape i\kern-0.25em b}\kern-0.8em\TeX}}}

\setcopyright{none}

\acmConference[DaSH@KDD]{}{August 24, 2020}{Virtual Conference}
\acmDOI{}
\acmISBN{}

\begin{document}

\title{Active Learning++: Incorporating Annotator's Rationale using Local Model Explanation}

\author{Bhavya Ghai}
\affiliation{%
  \institution{Stony Brook University}
  \city{Stony Brook}
  \country{USA}
}

\author{Q. Vera Liao, Yunfeng Zhang}
\affiliation{%
  \institution{IBM Research AI}
  \city{Yorktown Heights}
  \country{USA}}

\author{Klaus Mueller}
\affiliation{%
  \institution{Stony Brook University}
  \city{Stony Brook}
  \state{USA}
}

\renewcommand{\shortauthors}{Ghai, et al.}

\begin{abstract}
  We propose a new active learning (AL) framework, \textit{Active Learning++}, which can utilize an annotator's labels as well as it's rationale. Annotators can provide their rationale for choosing a label by ranking input features based on their importance for a given query. To incorporate this additional input, we modified the disagreement measure for a bagging-based Query by Committee (QBC) sampling strategy. Instead of weighing all committee models equally to select the next instance, we assign higher weight to the committee model with higher agreement with the annotator's ranking. Specifically, we generated a feature importance-based local explanation for each committee model.
  The similarity score between feature rankings provided by the annotator and the local model explanation is used to assign a weight to each corresponding committee model.
  This approach is applicable to any kind of ML model using model-agnostic techniques to generate local explanation such as LIME. 
  With a simulation study, we show that our framework significantly outperforms a QBC based vanilla AL framework.
\end{abstract}



\begin{teaserfigure}
  \centering
  \includegraphics[width=0.9\textwidth]{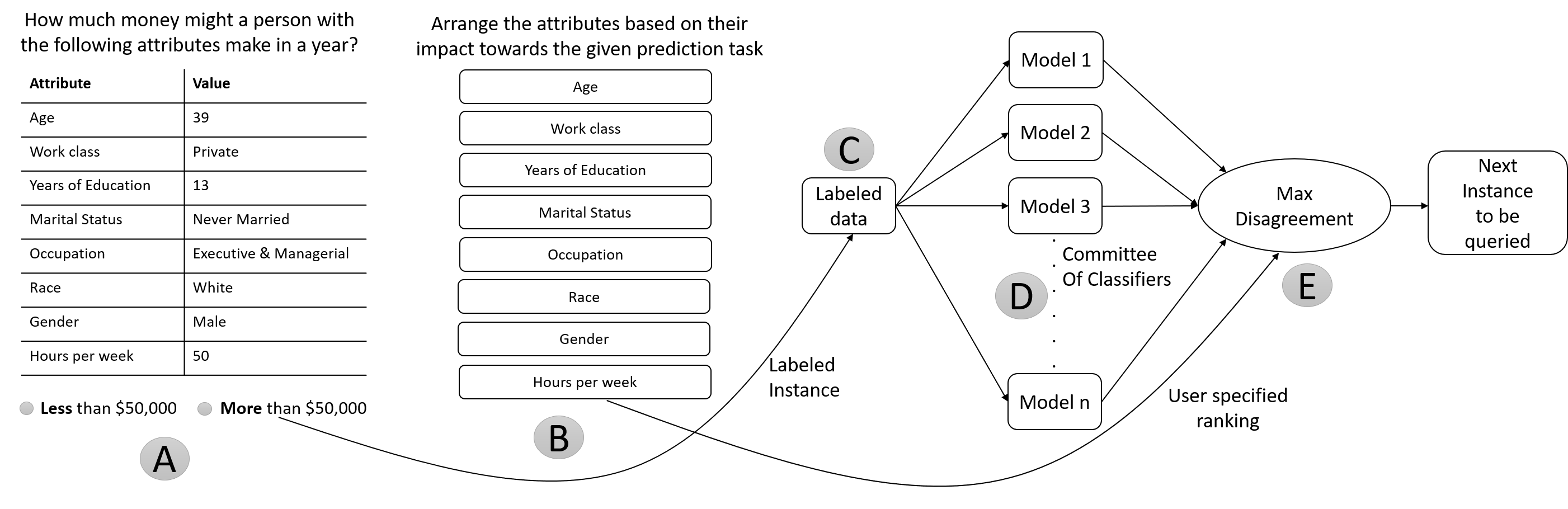}
  \caption{Workflow of our \textit{Active Learning++} framework. (A) represents a query and seeks its corresponding label. (B) elicits an annotator's rationale by arranging input features in decreasing order of importance. (C) represents the labeled dataset which is updated after each query. (D) represents the committee of classifiers used for the QBC sampling strategy. (E) represents the disagreement measure based on the model's prediction, local explanation and the annotator's rationale (as provided in (B)).  }
  \label{fig:teaser}
\end{teaserfigure}

\maketitle

\section{Introduction}
Active learning (AL) is a semi-supervised learning technique where the objective is to train a machine learning model using a minimal number of labeled training instances. Pool-based AL achieves this by intelligently selecting/sampling a batch of instances iteratively from a pool of unlabeled instances and getting them labeled by an oracle (human annotator) \cite{settles2009active}. The underlying premise is that some unlabeled instances are more informative than others and help train the ML model faster. This kind of learning technique plays a key role when labeled data is scarce and obtaining new labels is expensive or difficult. Some of the use cases include speech recognition, named entity recognition, text classification, etc. 

Typically, AL algorithms learn solely from the labels provided by the annotator. Our work relates more closely to the class of AL algorithms which queries feature-level input \cite{raghavan2007interactive,zaidan2007using}. The downside to such approaches is that it may be challenging for annotators, who are often not ML experts, to reason about all features of a model and provide robust input. Existing works are limited to text classification problems, where ``keywords'' based features are relatively intuitive to consider. We present a novel approach to elicit rationale in the form of feature ranking 
and incorporate this additional input as a weighing signal in the sampling strategy \cite{al_bagging}. This approach makes it easy for annotators to provide feature-level input, is relatively robust to partial or noisy input, and can be applied to problems beyond text classification.  We call this new AL framework \textit{Active learning++} as it can incorporate both instance labels and feature-level input. Our hypothesis is that richer input should help the model train faster than vanilla AL.

\section{Active Learning++}

We conducted an exploratory user study with an AL system\cite{ghai2020explainable} and found that annotators naturally want to ``teach'' with rationale beyond instance labels. One common type of rationale for justifying their choice of label is by attributing to the important contributing feature(s) of an instance. For example, one may cite job category and/or education as their rationale when asked to judge a person's income level. Hence, we elicit an annotator's rationale by asking them to rank input features based on their contribution in determining the label they give to the instance (see \autoref{fig:teaser}(B)).

To incorporate the rationale, we used a bagging based QBC sampling strategy \cite{al_bagging} and modified its disagreement measure. QBC uses a set of models (committee) to select the unlabeled instance where the predictions of individual committee members differs the most. 
There are multiple ways to quantify this disagreement, like voting, consensus entropy, etc.\cite{settles2009active}. In this work, we modified the \textit{Max Disagreement} measure to capture the disagreement such that models whose rationale is in higher agreement with the annotator's are given higher weight. 
To generate each model's rationale, we use an Explainable AI method called local feature importance, which characterizes the importance of each feature by its weight for the given prediction task. We then obtain a feature ranking for each model by sorting features based on their importance. Next, we quantify the agreement between a model's and the annotator's rationale by computing the similarity between the two feature rankings (rationales) using the Kendall's tau metric. Lastly, we factor in the similarity score for each classifier in the \textit{Max Disagreement} measure to fetch the next instance to be queried. Our approach is model agnostic as we can use any ML model to train the QBC committee of classifiers and then generate local explanations using model-agnostic techniques like LIME, SHAP, etc. This approach can be applied to any type of data with meaningful input features like text, tabular data, images, etc. 



\section{Methodology}
We simulated a pool-based AL pipeline with batch size 1 using the UCI Adult Census Income dataset \cite{adultUCI}. 
The prediction task was to classify the income of an individual given different attributes like Age, Education, Occupation, etc. into less/more than \$50,000 (see \autoref{fig:teaser} (A)). After data pre-processing, we divided the dataset into test and training at a 50:50 ratio. We randomly selected 5 instances from the training data for each output label which acted as our initial labeled dataset. From this pool of labeled instances, we trained 10 logistic regression models by randomly selecting 6 instances with replacement. These models acted as the committee members for the QBC sampling strategy. Furthermore, we used a logistic regression model which acted as the active learner.   

We compared our \textit{AL++} framework with a QBC based vanilla AL framework.  We made sure that all parameters such as the number of classifiers in the committee, initial labeled instances, etc. were identical for QBC and AL++. The only difference between the two is the disagreement measure (see \autoref{fig:teaser} (E)) where AL++ assign different weights to different classifiers. 

To simulate the oracle, one popular way is to simply use the ground-truth label. However,  AL++ would also require the ranking of input features. Hence, we trained a logistic regression model over the whole training dataset which can provide quality predictions (labels) along with their local explanation. Since it's a linear model, we simply multiplied the normalized feature values by the model coefficients to generate the local explanations. This model will act as the oracle for all sampling strategies. On the test dataset, this model had 84.22\% accuracy and a 0.65 F1 score. 

\section{Results}
\begin{figure}[tb]
 \centering 
 \includegraphics[width=\columnwidth]{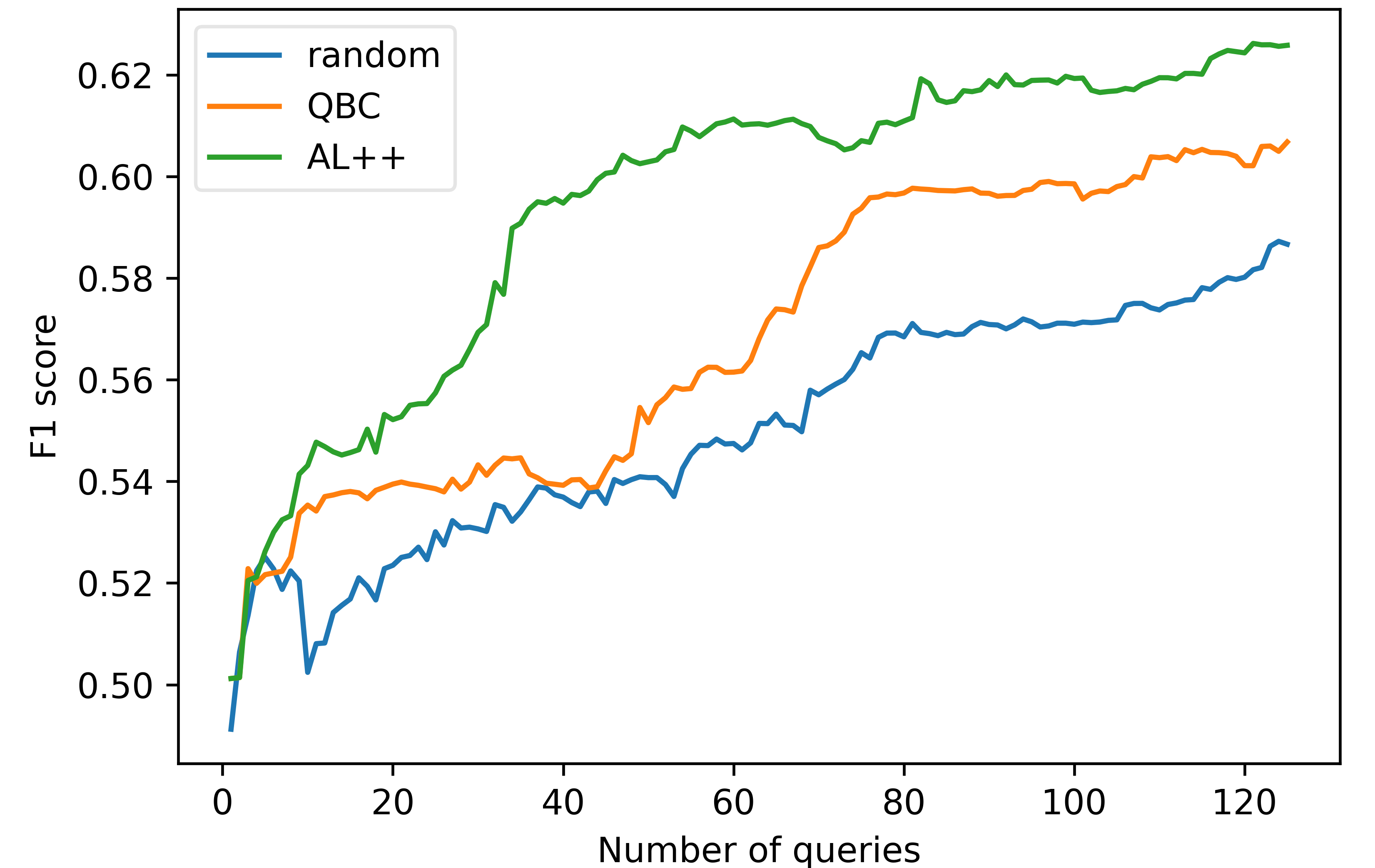}
 \caption{Comparing our proposed AL++ framework with with vanilla AL using Adult Income dataset. 
 }
 \label{fig:result}
\end{figure}
\autoref{fig:result} shows the F1 score for the random, QBC and AL++ strategies, respectively across 125 queries averaged over 10 experiments with different initial labeled datasets. We used the F1 metric over accuracy because the output class has an imbalanced distribution (ratio 25:75). From \autoref{fig:result}, we can clearly observe that AL++ maintains a constant lead over the other two strategies throughout. The mean F1 score across the 125 queries for AL++ (0.594) is much higher than for QBC (0.568). We found this difference to be statistically significant (paired t test, p<0.01). To reach a F1 score of 0.54, AL++ required 11.8 queries, much lower than QBC requiring 25.9 queries on average. 
It confirms our hypothesis that additional input in the form of feature ranking can help train a model faster. 

\section{Conclusion \& Future Work}
We presented a new AL framework, \textit{Active Learning++}, which can utilize annotator's rationale in the form of feature importance based ranking. 
Our simulation on the Adult Income dataset shows a significant performance jump over QBC based vanilla AL framework. 

Next, we plan to test our approach on different datasets and different domains such as text classification.
We also plan to conduct a user study to test this framework in practice. It will be interesting to see if our approach can help build fairer ML model given a fair oracle. Finally, we wish to simulate other interaction scenarios where the oracle provides rationale for only the top-k features. 


\begin{acks}
This research was partially supported by NSF grant IIS 1527200 and NSF grant IIS 1941613.
\end{acks}

\bibliographystyle{ACM-Reference-Format}
\bibliography{sample-base}


\end{document}